\documentclass[letterpaper, 10 pt, conference]{ieeeconf}  

\IEEEoverridecommandlockouts                              

\usepackage{algorithm}
\usepackage{algorithmic}
\usepackage{amsmath} 
\usepackage{bm}
\usepackage{amssymb}
\usepackage[dvipdfmx]{color}
\usepackage{cite}
\usepackage[dvipdfmx]{graphicx} 

\newcommand{\figref}[1]{{Fig.~\ref{#1}}}

\newcommand{\equref}[1]{{\eqref{#1}}} 

\newcommand{\refsec}[1]{Sec. \ref{sec:#1}}
\newcommand{\secref}[1]{\refsec{#1}} 

\newcommand{\scriptframe}[1]{\{#1\}}
\newcommand{\scriptsymbol}[1]{#1}

\newif\ifdraft
\draftfalse 

\newcommand{\revise}[1]{\textcolor{\ifdraft blue\else black\fi}{#1}}

\title{\LARGE \bf Singularity-free Aerial Deformation by Two-dimensional Multilinked Aerial Robot with 1-DoF Vectorable Propeller}

\author{Moju Zhao$^{1}$, Tomoki Anzai$^{1}$, Kei Okada$^{1}$, Masayuki Inaba$^{1}$
\thanks{$^{1}$
  M. Zhao, K. Okada and M. Inaba are with Department of Mechano-Infomatics, The University of Tokyo, 7-3-1 Hongo, Bunkyo-ku, Tokyo 113-8656, Japan
{\tt\small chou@jsk.t.u-tokyo.ac.jp}}
}

\begin{document}

\maketitle
\thispagestyle{empty}
\pagestyle{empty}

\begin{abstract}
\revise{Two-dimensional multilinked structures can benefit aerial robots in both maneuvering and manipulation because of their deformation ability. However, certain types of singular forms must be avoided during deformation. Hence, an additional 1 Degrees-of-Freedom (DoF) vectorable propeller is employed in this work to overcome singular forms by properly changing the thrust direction.}
\revise{In this paper, we first extend modeling and control methods from our previous works for an under-actuated model whose thrust forces are not unidirectional. We then propose a planning method for the vectoring angles to solve the singularity by maximizing the controllability under arbitrary robot forms.}
  Finally, we demonstrate the feasibility of the proposed methods by experiments where a quad-type model is used to perform  \revise{trajectory tracking} under challenging forms, such as a line-shape form, and the deformation passing these challenging forms.

\end{abstract}

\section{Introduction}
\label{sec:intro}

\begin{figure}[t]
  \begin{center}
    \includegraphics[width=\columnwidth]{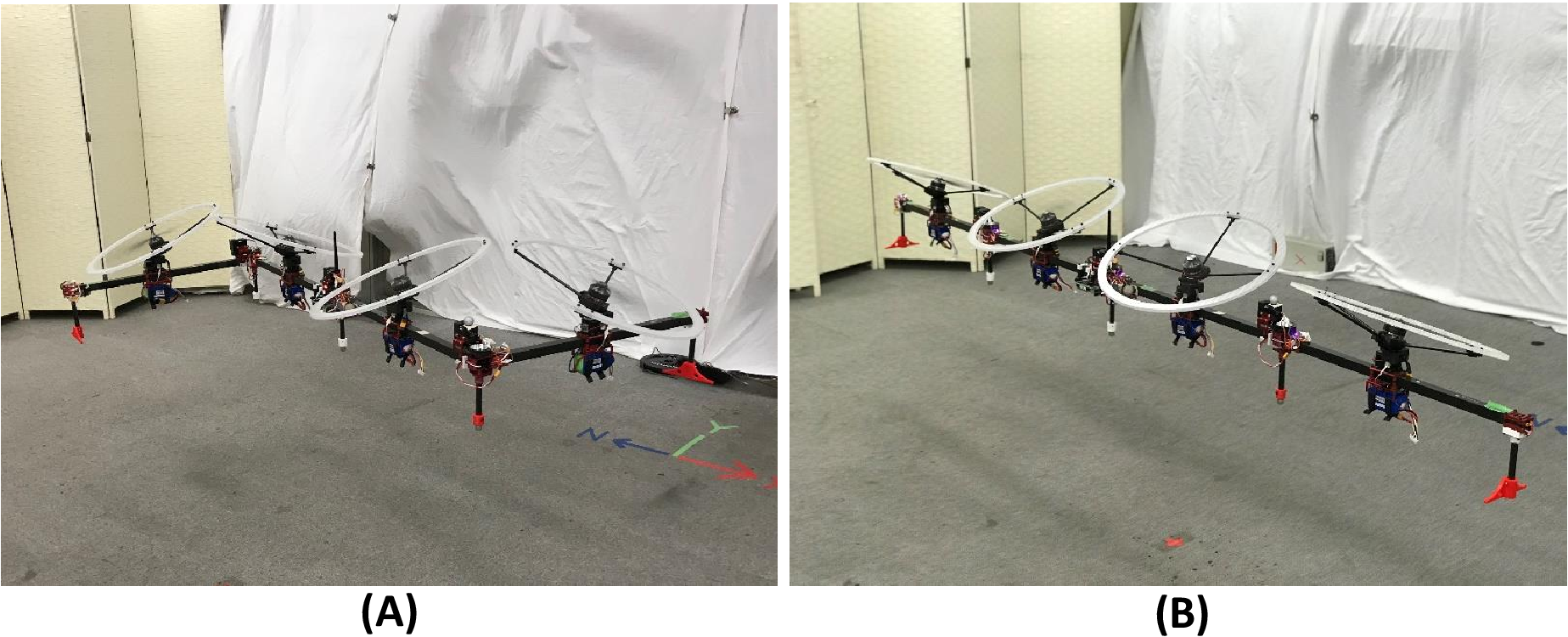}
    \vspace{-6mm}
    \caption{Stable flight achieved by a two-dimensional multilinked aerial robot with \revise{1-DoF yaw-vectorable propeller} based on the proposed modeling, control, and planning methods. Without the vectorable propeller, these two forms would be singular, since the rotational controllability would entirely vanish, especially under the form of (B), where all rotors are aligned along one straight line.}
    \label{figure:abst}
    \vspace{-5mm}
  \end{center}
\end{figure}

\revise{Compared with traditional multirotors of the same size, aerial deformability can benefit the aerial robot not only in maneuvering in confined environments \cite{soft-robotics-deformable, iros2017-deformable, zurich-deformable}, but also in aerial manipulation \cite{upenn-deformable-gripper-icra2018, odar-lasdra-manipulation-icra2018}}. Among various deformable designs, the two-dimensional multilinked structure proposed in our previous work \cite{hydrus-ijrr2018} demonstrates advanced performance in both maneuvering and manipulation.
However, the main issue of such a multilinked model is \revise{the singularity where the controllability of one of three rotational Degrees-of-Freedom (DoF) is lost with certain joint configurations, thus preventing stable hover control}. One typical singular form is the line-shape form, where all rotors are aligned in the same straight line, leading to zero rotational moment around this line. \revise{Although the joint trajectory planning method proposed by \cite{hydrus-ar2016} can avoid singular forms during deformation, advanced maneuvering, such as flight through long and narrow corridors by deforming into a line form, cannot be achieved.}
\revise{We found that a net force with variable direction in a three-dimensional force space is the key to overcoming the singularity.}
\revise{Then, \cite{anzai-hydrus-xi-iros2019} proposed a vectoring apparatus for propellers that can generate a lateral component force with variable direction; they also developed a prototype composed of six links embedded with this propeller apparatus to achieve flight stability in a fully-actuated manner. However, the singularity issue is not addressed in this previous work. Furthermore, the feasibility of such a vectoring apparatus for model with less than six links, which is considered under-actuated for fixed joint angles and fixed propeller vectoring angles, has not been evaluated. Thus, in the current study, fully singularity-free aerial deformation, especially for such an under-actuated model as shown in \figref{figure:abst}, is investigated from the aspects of modeling, control, and planning.}

In terms of the propeller vectoring apparatus used  \revise{to generate a variable net force},
most of the proposed designs allow a rotor rotating around an axis perpendicular to the rotor shaft \cite{bi-copter-iros2015, tohoku-vectoring-quadrotor-icra2015, eth-voliro-ram2018}; \revise{this can be  defined as a roll-vectoring apparatus. However, a crucial shortcoming of such an apparatus is that it cannot generate  lateral force in arbitrary directions. In contrast, the apparatus proposed by\cite{anzai-hydrus-xi-iros2019}, named a yaw-vectoring apparatus, has a vectoring axis perpendicular to the plane of the two-dimensional multilinked model and also has a rotor mount tilted from the vectoring axis at a fixed angle to yield the lateral component force.}
\revise{It is notable that, the vectoring angles cannot be easily used as a control input, because the opposite lateral component force demands the apparatus to rotate 180$^{\circ}$ back and forth at a high rate, thereby making the flight significantly unstable.}
\revise{Instead, these vectoring angles are considered configuration parameters like the joint angles.}

Regarding modeling and control, \revise{we treat the rotor speed (i.e., thrust force) as the only control input;
  however, tilted rotor mounts (and thus the propellers) indicate a full influence on the wrench space.}
The influence of such tilting propellers on under-actuated models (e.g., four rotors) is more challenging than that on  fully-actuated models \cite{se3-hexrotor-control, eth-odar-icra2016, odar-tasme2018}. Although \cite{shi-tilt-hydrus-mpc-ar2019} proposed a nonlinear model predictive control method for under-actuated models, uncertain model error is not allowed; consequently, robustness under singular forms such as that in \figref{figure:abst} is weak. \revise{Thus, modeling under near-hover assumption and the cascaded control framework proposed by \cite{hydrus-ijrr2018} are applied in this work. However, a proper orientation should be obtained for center-of-gravity (CoG) frame, which is always level in hover state. In addition, the undesired lateral component force from the attitude control should be suppressed, since the horizontally translational motion should be controlled by the rotational motion.}

In terms of planning for vectoring angles, \cite{anzai-hydrus-xi-iros2019} proposed a method that jointly solves trajectories for the joint and vectoring angles. The strength of this planning method is its ability to find an effective joint trajectory between a start and goal forms, thereby avoiding singular forms. However, the multilinked model in this work should be \revise{singularity-free}, which means that planning vectoring angles for arbitrary forms is more important than planning joint trajectory. Hence, \revise{a more robust planning method based on complexity reduction for vectoring angles} is developed in this work to enable instant response to  changes in joint angles.

\revise{The main contributions of this work compared with our previous work \cite{hydrus-ijrr2018, hydrus-ar2016, anzai-hydrus-xi-iros2019}  can be summarized as follows:
  \begin{enumerate}
  \item  We present a modeling method for obtaining the hovering thrust and a proper CoG frame  for under-actuated models with yaw-vectorable propellers.
  \item  We extend the cascaded control framework by adding an additional cost term in the LQI based attitude control to suppress the undesired lateral force.
  \item  We develop a real-time optimization-based  configuration planner for vectoring angles to achieve singularity-free flight and deformation.
  \end{enumerate}
}

The remainder of this paper is organized as follows. The design of the proposed multilinked model is presented in \secref{design}. The modeling and control methods are presented in \secref{modeling} and  \secref{control}, respectively, followed by the planning method for vectoring angles in \secref{planning}. We show the experimental results in \secref{experiment} before concluding in \secref{conclusion}.

\section{Design}
\label{sec:design}
\begin{figure}[b]
  \begin{center}
    \includegraphics[width=\columnwidth]{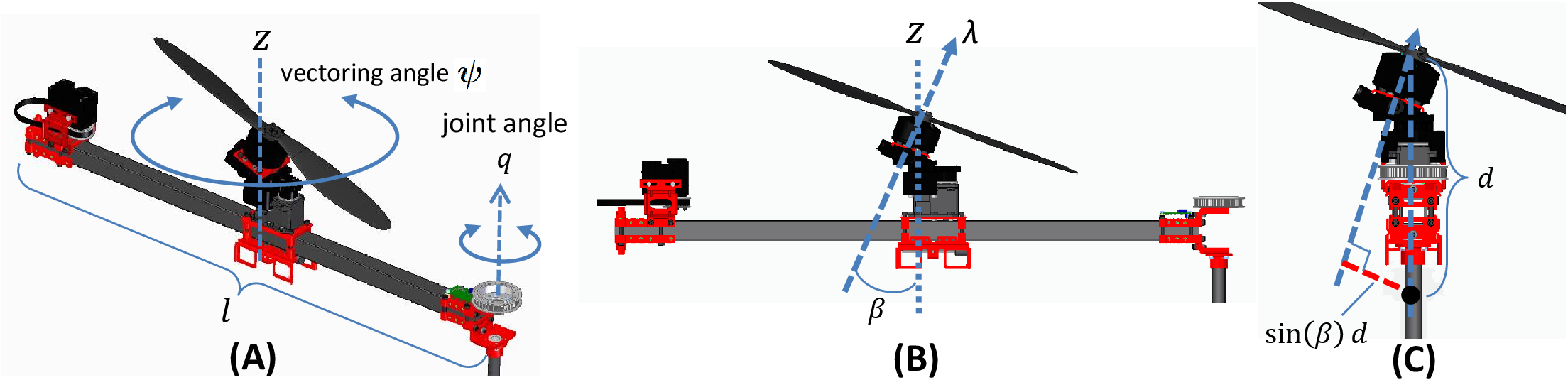}
    \vspace{-5mm}
    \caption{{\bf A}: Structure of the link module consisting of a propeller yaw-vectoring apparatus and link joints at each end.\revise{ $\psi$ is the vectoring angle around the $z$ axis, and $q$ is the joint angle.}  {\bf B}: The rotor and propeller are tilted from the $z$ axis at a fixed angle of $\beta$, and $\lambda$ denotes the thrust magnitude.
      {\bf C}:  Even when all links are aligned in the same straight line, the moment around this line (which is equal to $sin(\beta) \lambda d$) can still be generated to  avoid the control singularity under this form. $d$ is the distance from the propeller to the origin of the robot CoG.}
    \vspace{-3mm}
    \label{figure:design}
  \end{center}
\end{figure}

\subsection{Link Module}
The link module, as shown in \figref{figure:design}(A), contains a joint part actuated by a servo motor at each end. All joint axes are parallel, resulting in two-dimensional deformation.

On the other hand, there is a 1 DoF yaw-vectoring apparatus at the center of the link module which employs another servo motor under the rotor mount to rotate it around the $z$ axis as shown in \figref{figure:design}(A), and we define the rotating (vectoring) angle as $\psi$.
Given that $\psi$ is considered a configuration parameter but not a control input, a relatively  small servo motor without strong torque output can be used. Thus the weight increase can be suppressed to some degree \revise{compared with a model without such a vectoring apparatus.}
Furthermore, there is a fixed tilting angle $\beta$ between the rotor-rotating axis and the $z$ axis of the link module as shown in \figref{figure:design}(B) to \revise{enable a net force with variable direction in the three-dimensional force space.}

\subsection{Decision on Propeller Tilting Angle}
\revise{
The propeller tilting angle $\beta$ divides the thrust force into the vertical component force $cos(\beta)\lambda$ and the lateral component force $sin(\beta)\lambda$. The former force is mainly used to balance with gravity, which is related to the energy efficiency, whereas the latter force has three aspects. First, it enhances the overall yaw torque for all forms. Second, under the line-shape singular form as shown in \figref{figure:abst}(B), it guarantees the torque around the line with proper vectoring angles that make thrust forces perpendicular to the line. Third, it affects the horizontally translational motion, which is considered a disturbance in the under-actuated model.
Given that the first aspect has a positive correlation with the second aspect, we mainly focus on the second and third aspects. It is also notable that a smaller $\beta$ is preferred regarding the third aspect. Then, we design the following optimization objective to minimize $\beta$ (i.e., the third aspect) with the consideration of the energy efficiency and the guaranteed torque under the line-shape form  (i.e., the second aspect):
\begin{eqnarray}
  \label{eq:optimization_beta}
  & \displaystyle \min \beta, \\
  \label{eq:optimization_beta_cons}
  & s.t. \ \frac{1}{cos(\beta)}  \leq \gamma_1,  \  \frac{4sin(\beta)d}{l} \geq \gamma_2.
\end{eqnarray}
The first constraint term of \equref{eq:optimization_beta_cons} denotes  the energy efficiency, since the total thrust force needed to balance with gravity is $\sum \lambda_i \approx \frac{m_{\Sigma}g}{cos{\beta}}$, where $m_{\Sigma}$ is the total mass. The second constraint term is the ratio between the maximum torque $2sin(\beta)\lambda_{\mathrm{max}}d$ around the singular line under the line-shape form, and the approximated maximum torque  $\frac{\lambda_{\mathrm{max}}l}{2}$ around the CoG under the normal form (i.e., joint angle $q_i = 90^{\circ}$ for the four link model). $l$ is the link length and is set as 0.6 m in our model. $d$ is the distance from the CoG to the propeller, as shown in \figref{figure:design}(C), which is also critical for generating the torque. Given that all batteries are attached under the link rod, the CoG is relatively low, and sufficient $d$ (i.e., 0.1 m) is guaranteed.
}

\revise{
Finally, we obtain the result of $\beta = 0.34$ rad $\approx 20^{\circ}$ with  $\gamma_1 = 1.05$ and $\gamma_2 = 0.2$. This result indicates that a 5\% increase in thrust force is required to balance with gravity compared with the model without a tiling angle, which can however be considered relatively small and neglected.
}



\section{Modeling}
\label{sec:modeling}

\subsection{Notation}
\revise{From this section, boldface symbols (e.g., $\bm{r}$) denote vectors, whereas non-boldface symbols (e.g., $m$ or $I$) denote scalars or matrices.
The coordinate system in which a vector is expressed is denoted by a superscript, e.g., ${}^{\scriptframe{W}}\bm{r}$ expressing $\bm{r}$ with reference to (w.r.t.) the frame $\{W\}$. Subscripts are used to express a target frame or an axis, e.g., ${}^{\scriptframe{W}}\bm{p}_{\scriptframe{B}}$ representing a vector point from $\{W\}$ to $\{B\}$ w.r.t. $\{W\}$, whereas $p_x$ denotes the $x$ component of the vector $\bm{p}$.}

\subsection{Wrench Allocation}
\revise{
As shown in \figref{figure:kinematics}, the wrench generated by each rotor w.r.t. the frame $\{CoG\}$ can be written as follows:
\begin{align}
\label{eq:wrench_force}
{}^{\scriptframe{CoG}}\bm{f}_{i} &= \lambda_i {}^{\scriptframe{CoG}}R_{\scriptframe{F_i}}(\bm{q}, \bm{\psi}) \bm{b}_3, \\
\label{eq:wrench_torque}
      {}^{\scriptframe{CoG}}\bm{\tau}_{i} &=  {}^{\scriptframe{CoG}}\bm{p}_{\scriptframe{F_i}}(\bm{q}, \bm{\psi}) \times {}^{\scriptframe{CoG}}\bm{f}_{i} + \kappa_i  {}^{\scriptframe{CoG}}\bm{f}_{i}, \nonumber \\
      & = \lambda_i ({}^{\scriptframe{CoG}}\hat{\bm{p}}_{\scriptframe{F_i}}(\bm{q}, \bm{\psi}) \\
      &+ \kappa_i E_{3 \times 3} ) {}^{\scriptframe{CoG}}R_{\scriptframe{F_i}}(\bm{q}, \bm{\psi}) \bm{b}_3,
\end{align}
where  $i = \{1, 2, 3, 4\}$. $\bm{q} \in \mathcal{R}^3$ and $\bm{\psi}  \in \mathcal{R}^4$ denote the joint and vectoring angles, respectively. These variables influence ${}^{\scriptframe{CoG}}R_{\scriptframe{F_i}}(\bm{q}, \bm{\psi})$ and ${}^{\scriptframe{CoG}}{\bm p}_{\scriptframe{F_i}}(\bm{q}, \bm{\psi})$, which are the orientation and the origin position of the rotor frame $\{F_i\}$ w.r.t the frame $\{CoG\}$. Note that the rotor frame $\{F_i\}$ is attached to the rotor mount, and the thrust force  $\lambda_i$ is along the $z$ axis of $\{F_i\}$.  $\bm{b}_3$ is a unit vector $\left[\begin{array}{ccc}0 & 0& 1 \end{array} \right]^{\mathrm{T}}$ and $\kappa_i$ is the rate regarding the rotor drag moment. $\hat{\cdot}$ denotes the operation from a vector to a skew-symmetric matrix, while $E_{3 \times 3}$ is a 3$\times$ 3 identity matrix.}

\revise{
Using \equref{eq:wrench_force} and \equref{eq:wrench_torque}, allocation from  the thrust force $\bm{\lambda} = \left[\lambda_1 \ \lambda_2 \ \lambda_3 \ \lambda_4 \right]^{\mathrm{T}} $ to the total wrench can be given by:
\vspace{-1mm}
\begin{eqnarray}
\label{eq:total_wrench1}
\left(
\begin{array}{c}
  {}^{\scriptframe{CoG}}\bm{f} \\
  {}^{\scriptframe{CoG}}\bm{\tau} \\
\end{array}
\right)
=
{\displaystyle \sum_{i = 1}^{4}} \left(
\begin{array}{c}
  {}^{\scriptframe{CoG}}\bm{f}_i \\
  {}^{\scriptframe{CoG}}\bm{\tau}_i \\
\end{array}
\right)
=
Q(\bm{q}, \bm{\psi}) \bm{\lambda},
\end{eqnarray}
\vspace{-1mm}
where  $Q(\bm{q}, \bm{\psi}) \in \mathcal{R}^{6 \times 4} $.}

\subsection{Definition of CoG Frame}
\revise{
The origin of the frame $\{CoG\}$ w.r.t the root link frame $\{L_1\}$  can be straightforwardly written as:
\begin{eqnarray}
\label{eq:cog_oirigin}
      {}^{\scriptframe{L_1}}\bm{p}_{\scriptframe{CoG}} &=& \frac{1}{{m_{\Sigma}}}\displaystyle \sum_i^4 m_{L_i} {}^{\scriptframe{L_1}}\bm{p}_{C_i}(\bm{q}, \bm{\psi}),
\end{eqnarray}
where  $m_{\scriptsymbol{L_i}}$ is the mass of the $i$-th link (i.e., ${m_{\Sigma}} = \sum m_{\scriptsymbol{L_i}}$). ${}^{\scriptframe{L_1}}\bm{p}_{C_i}(\bm{q}, \bm{\psi})$ is the position of the center-of-mass of the $i$-th link (i.e., $C_i$) w.r.t. the root link frame $\{L_1\}$.}

\revise{
Regarding the frame orientation, given the near-hover assumption in the control method presented in \secref{control}, the frame $\{CoG\}$ is required to be level (horizontal) in the ideal hover state. Mathematically, the frame $\{CoG\}$ should satisfy the following equations to find the static thrust force $\bm{\lambda}_s$:
\begin{eqnarray}
  \label{eq:static_thrust_force1}
  Q_{\mathrm{t}}(\bm{q}, \bm{\psi}) \bm{\lambda}_s &=& m_{\Sigma}\bm{g}, \\
  \label{eq:static_thrust_torque1}
 Q_{\mathrm{r}}(\bm{q}, \bm{\psi}) \bm{\lambda}_s &=& \bm{0},
 \end{eqnarray}
where 
\begin{eqnarray*}
  \left[\begin{array}{c} Q_{\mathrm{t}}(\bm{q}, \bm{\psi}) \\ Q_{\mathrm{r}}(\bm{q}, \bm{\psi})\end{array}\right] = Q(\bm{q}, \bm{\psi}) ; Q_{\mathrm{t}}(\bm{q}, \bm{\psi}), Q_{\mathrm{r}}(\bm{q}, \bm{\psi}) \in \mathcal{R}^{3\times 4},
\end{eqnarray*}
 $\bm{g} = \left[0 \ 0 \ g\right]^{\mathrm{T}}$ is aligned on the $z$ axis of the frame $\{W\}$.
}

\revise{
For the multilinked model in \cite{hydrus-ijrr2018, hydrus-ar2016, anzai-hydrus-xi-iros2019}, the $z$ axis of the frame $\{CoG\}$ is set to be perpendicular to the two-dimensional multilinked plane, since the thrust forces are unidirectional, which satisfies \equref{eq:static_thrust_force1} and \equref{eq:static_thrust_torque1}. Then the direction of the $x$ and $y$ axes is usually set to be identical with that of the root link $\{L_1\}$. However, for the proposed model in this work, the thrust force is not unidirectional, and such an orientation of $\{CoG\}$ can hardly satisfy \equref{eq:static_thrust_force1} since $Q_{\mathrm{t}}(\bm{q}, \bm{\alpha})$ has full influence on the force space.
Nevertheless, there must exist a rotation matrix $R$ for $Q_{\mathrm{t}}(\bm{q}, \bm{\psi})$ and $Q_{\mathrm{r}}(\bm{q}, \bm{\psi})$ that satisfies $RQ_{\mathrm{t}}(\bm{q}, \bm{\psi}) \bm{\lambda}_s = m_{\Sigma}\bm{g}, RQ_{\mathrm{r}}(\bm{q}, \bm{\psi}) \bm{\lambda}_s = \bm{0}$. 
Thus, $R$ is the key to defining the orientation of $\{CoG\}$ for our model.}

\begin{figure}[t]
  \begin{center}
    \includegraphics[width=\columnwidth]{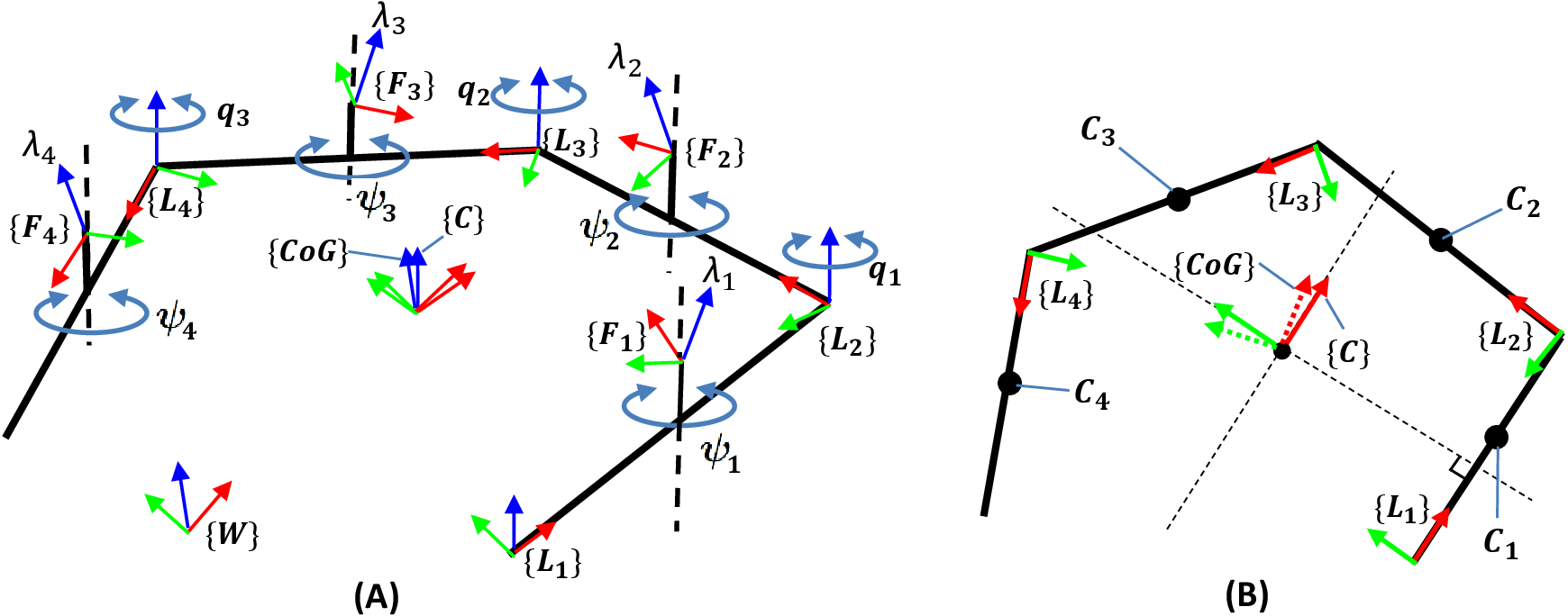}
    \vspace{-6mm}
    \caption{\revise{Kinematics model of the multilinked structure from the perspective view (A) and top view (B). The frame $\{W\}$ is an inertial reference frame where the gravity is along the $z$ axis. The frame $\{C\}$ and $\{CoG\}$ have the same origin at the center-of-mass of the whole model. The orientation of the frame $\{C\}$ is identical to the frame $\{L_1\}$, as shown in (B), while the orientation of the frame $\{CoG\}$ is specially designed as discussed in \secref{modeling}. C. The frame $\{F_i\}$ is attached to each rotor mount, and the thrust force $\lambda_i$ yields along the $z$ axis of $\{F_i\}$ as shown in (A). $\bm{C}_i$ in (B) is the center-of-mass of the $i$-th link module. }}
    \vspace{-5mm}
    \label{figure:kinematics}
  \end{center}
\end{figure}

\revise{
  Now, we assume that the correct orientation of $\{CoG\}$ is unknown.
  Here we introduce a candidate frame $\{C\}$ that has the same orientation as that of $\{L_1\}$ and also satisfies \equref{eq:cog_oirigin} as shown in \figref{figure:kinematics}. Then, by deriving \equref{eq:wrench_force} $\sim$ \equref{eq:total_wrench1} with $\{C\}$ instead of $\{CoG\}$, we can obtain different allocation matrices $Q^{'}_{\mathrm{t}}(\bm{q}, \bm{\psi})$ and $Q^{'}_{\mathrm{r}}(\bm{q}, \bm{\psi})$. Furthermore, we relax the constraints of \equref{eq:static_thrust_force1} and \equref{eq:static_thrust_torque1} for $\{C\}$ by only focusing on the $z$ axis force and three-dimensional torque as follows:
\begin{equation}
  \label{eq:lambda_dash}
\left(
  \begin{array}{c}
    Q^{'}_{\mathrm{t_z}}(\bm{q}, \bm{\psi}) \\
    Q^{'}_{\mathrm{r}}(\bm{q}, \bm{\psi}) \\
  \end{array}
  \right) \bm{\lambda}^{'} = \left[\begin{array}{cccc} 1 & 0 & 0 &0 \end{array} \right]^{\mathrm{T}},
\end{equation}
where 
$Q^{'}_{\mathrm{t_z}}(\bm{q}, \bm{\psi})\in\mathcal{R}^{1 \times 4}$ is the third row vector of $Q^{'}_{\mathrm{t}}(\bm{q}, \bm{\psi})$. Note that, $\bm{\lambda}^{'}$ can be easily obtained from \equref{eq:lambda_dash} by calculating the inverse matrix of the matrix in the left side. The first element in the right side of \equref{eq:lambda_dash} denotes 1 N on the $z$ axis of the force $Q^{'}_{\mathrm{t}}(\bm{q}, \bm{\psi}) \bm{\lambda}^{'}$, and there must exist  non zero values in both the $x$ and $y$ axes. Then through calculation of the norm  $\|Q^{'}_{\mathrm{t}}(\bm{q}, \bm{\psi}) \bm{\lambda}^{'}\|$, the static thrust $\bm{\lambda}_s$ can be derived by:
\begin{eqnarray}
  \label{eq:lambda_s}
  \bm{\lambda}_s &=& \frac{m_{\Sigma}g}{\|Q^{'}_{\mathrm{t}}(\bm{q}, \bm{\psi}) \bm{\lambda}^{'} \|} \bm{\lambda}^{'}.
\end{eqnarray}
}

\revise{
Further, the rotation matrix between the frame $\{C\}$ and the target frame $\{CoG\}$ can be written by:
\begin{eqnarray}
  \label{eq:R_CoG}
{}^{\scriptframe{CoG}}R_{\scriptframe{C}} Q^{'}_{\mathrm{t}}(\bm{q}, \bm{\psi}) \bm{\lambda}_s =  m_{\Sigma}\bm{g}.
\end{eqnarray}}

\revise{
By introducing Euler angles $\bm{\alpha} = \left[\alpha_x \ \alpha_y \ \alpha_z \right]^{\mathrm{T}}$ and $\left[f_x \ f_y \ f_z\right]^{\mathrm{T}} = Q^{'}_{\mathrm{t}}(\bm{q}, \bm{\psi}) \bm{\lambda}_s$, ${}^{\scriptframe{CoG}}R_{\scriptframe{C}}$  can be further given by:
\begin{eqnarray}
  \label{eq:R_CoG2}
        {}^{\scriptframe{CoG}}R_{\scriptframe{C}} &=& R_Y(\alpha_y) R_{\revise{X}}(\alpha_x), \\
        \label{eq:R_CoG_roll}
        \alpha_x &=& tan^{-1}(f_{y}, f_{z}), \\
        \label{eq:R_CoG_pitch}
        \alpha_y &=& tan^{-1}(-f_{x}, \sqrt{f_{y}^2 + f_{z}^2}),
\end{eqnarray}
where  $R_{X}(\cdot), R_{Y}(\cdot)$ denote the rotation matrix, which only rotates along the $x$ and $y$ axis, respectively, with certain angles.}

\revise{
In summary, the position and the orientation of the frame $\{CoG\}$ can be obtained from \equref{eq:cog_oirigin} and \equref{eq:R_CoG2} respectively. It is notable that \equref{eq:R_CoG2} reveals that the $z$ axis of the  frame $\{CoG\}$ is not necessarily perpendicular to the two-dimensional multilinked plane in most of case as shown in \figref{figure:kinematics}, and thus the two-dimensional multilinked plane is not necessarily level in the ideal hover state.
}

\revise{
Finally, the allocation matrix $Q(\bm{q}, \bm{\psi})$ in \equref{eq:total_wrench1} can be rewritten by:
\begin{eqnarray}
  \label{eq:total_wrench}
Q(\bm{q}, \bm{\psi})
&=& \left(
\begin{array}{c}
  Q_{\mathrm{t}}(\bm{q}, \bm{\psi}) \\
  Q_{\mathrm{r}}(\bm{q}, \bm{\psi}) \\
\end{array}
\right)
\nonumber \\
&=& \left(
\begin{array}{c}
  {}^{\scriptframe{CoG}}R_{\scriptframe{CoG}}  Q^{'}_{\mathrm{t}}(\bm{q}, \bm{\psi}) \\
{}^{\scriptframe{CoG}}R_{\scriptframe{CoG}}  Q^{'}_{\mathrm{r}}(\bm{q}, \bm{\psi}) \\
\end{array}
\right). 
\end{eqnarray}
}

\section{Control}
\label{sec:control}
\begin{figure}[b]
  \begin{center}
    \includegraphics[width=\columnwidth]{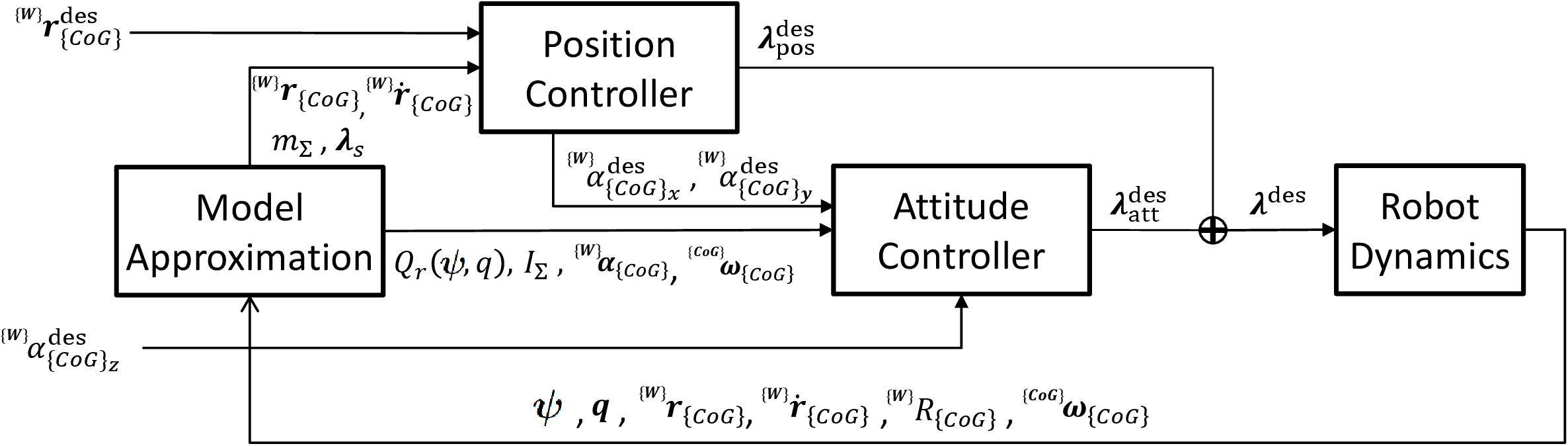}
    \vspace{-5mm}
    \caption{Proposed control framework for multilinked aerial robot. This framework involves dynamics model approximation  and a cascaded control flow, including position and attitude controllers. \revise{The reference of this cascaded control flow is the desired 3D position ${}^{\scriptframe{W}}\bm r^{\scriptsymbol{\mathrm{des}}}_{\scriptframe{CoG}}$ and the desired yaw angle ${}^{\scriptframe{W}} \alpha^{\scriptsymbol{\mathrm{des}}}_{\scriptframe{CoG}_z}$. The inputs are the position ${}^{\scriptframe{W}}\bm r_{\scriptframe{CoG}}$, velocity ${}^{\scriptframe{W}}\bm r_{\scriptframe{CoG}}$, Euler angles ${}^{\scriptframe{W}}\bm {\alpha}_{\scriptframe{CoG}}$, angular velocity ${}^{\scriptframe{CoG}}\bm {\omega}_{\scriptframe{CoG}}$, inertial parameters $m_{\scriptsymbol{\sum}}, I_{\scriptsymbol{\sum}}$, static thrust $\bm{\lambda}_s$, and the allocation matrix  $Q_r(\bm{q}, \bm{\psi})$; the output is the  desired thrust $\bm{\lambda}^{\scriptsymbol{\mathrm{des}}}$. }}
    \label{figure:control_diagram}
        \vspace{-5mm}
  \end{center}
\end{figure}

\revise{The control framework in this work is similar to that in our previous study \cite{hydrus-ijrr2018}, which contains a  model approximation followed by a cascaded control flow as shown in \figref{figure:control_diagram}. The main difference between the previous and current works is the extended cost function for optimal control in the attitude controller of this study.}

\subsection{Approximated Dynamics}
\revise{In this work, the multilinked model is regarded as a time-variant rigid body, since the joint motion is assumed to be sufficiently slow, as was assumed in our previous work \cite{hydrus-ar2016}}. Thus, the dynamics regarding the frame $\{CoG\}$ can be simplified as follows:
\begin{align}
  \label{eq:translational_dynamics}
&  m_{\sum} ({}^{\scriptframe{W}}\ddot{\bm r}_{\scriptframe{CoG}} + \bm g) = {}^{\scriptframe{W}}R_{\scriptframe{CoG}} Q_{\mathrm{t}}(\bm{q}, \bm{\psi}) \bm{\lambda}, \\
  \label{eq:rotational_dynamics}
& {}^{\scriptframe{CoG}}{\bm \omega}_{\scriptframe{CoG}} \times  {}^{\scriptframe{CoG}}I_{\sum}(\bm{q}, \bm{\psi})  {}^{\scriptframe{CoG}}{\bm \omega}_{\scriptframe{CoG}} \nonumber \\ & + {}^{\scriptframe{CoG}}I_{\sum}(\bm{q}, \bm{\psi})  {}^{\scriptframe{CoG}}\dot{\bm \omega}_{\scriptframe{CoG}} =  Q_{\mathrm{r}}(\bm{q}, \bm{\psi}) \bm{\lambda}, 
\end{align}
where \revise{$Q_{\mathrm{t}}(\bm{q}, \bm{\psi})$ and $Q_{\mathrm{r}}(\bm{q}, \bm{\psi})$ are the allocation matrices from \equref{eq:total_wrench}.}
${}^{\scriptframe{W}}\bm r_{\scriptframe{CoG}}$ and ${}^{\scriptframe{W}}R_{\scriptframe{CoG}}$ are the position and orientation, respectively, of the frame $\{CoG\}$  w.r.t. the world frame $\{W\}$.
These states can be calculated based on the forward kinematics from the states of the root link (i.e., ${}^{\scriptframe{W}}\bm r_{\{L_1\}}$, ${}^{\scriptframe{W}}\dot{\bm r}_{\{L_1\}}$, and ${}^{\scriptframe{W}}R_{\{L_1\}}$), while the angular velocity \revise{${}^{\scriptframe{CoG}}\bm \omega_{\scriptframe{CoG}}$} can be obtained by $\revise{{}^{\scriptframe{CoG}}\bm \omega_{\scriptframe{CoG}}} = {}^{\scriptframe{CoG}}R_{\{C\}}  {}^{\{L_1\}}{\bm \omega}_{\{L_1\}}$.
The total inertial matrix ${}^{\scriptframe{CoG}}I_{\sum}(\bm{q}, \bm{\psi})$ w.r.t. the frame $\{CoG\}$ can also be calculated from the forward-kinematics process.

Further, we assume that most of flight tasks are performed under near-hover condition. Thus, the approximation between the differential of the Euler angles and angular velocity $ {}^{\scriptframe{W}}\dot{\bm \alpha}_{\scriptframe{CoG}} \approx {}^{\scriptframe{CoG}}{\bm \omega}_{\scriptframe{CoG}}$ is available, since ${}^{\scriptframe{W}}\alpha_{\scriptframe{CoG}_x} \approx 0$, ${}^{\scriptframe{W}}\alpha_{\scriptframe{CoG}_y} \approx 0$. Then, the rotational dynamics expressed in \equref{eq:rotational_dynamics} can be further linearized under the near-hover condition:
\begin{eqnarray}
  \label{eq:rotational_dynamics2}
 {}^{\scriptframe{CoG}}{\bm \omega}_{\scriptframe{CoG}} \times  {}^{\scriptframe{CoG}}I_{\sum}(\bm{q}, \bm{\psi}) {}^{\scriptframe{CoG}}{\bm \omega}_{\scriptframe{CoG}} \nonumber  \\ + {}^{\scriptframe{CoG}}I_{\sum}(\bm{q}, \bm{\psi})  {}^{\scriptframe{W}}\ddot{\bm \alpha}_{\scriptframe{CoG}} = Q_{\mathrm{r}}(\bm{q}, \bm{\psi}) \bm{\lambda}.
\end{eqnarray}
Finally, \equref{eq:translational_dynamics} and \equref{eq:rotational_dynamics2} are the basic dynamics used in the proposed control framework.
\revise{It is also notable that $Q_{\mathrm{t}}(\bm{q}, \bm{\psi}) \bm{\lambda}$  in \equref{eq:translational_dynamics} has the full influence on the 3D force space, which is the main difference from a unidirectional multirotor, but is also the main issue to be solved in our control method.}

\subsection{Attitude Control}
\revise{Given that it is necessary to suppress the lateral force from the attitude control part in the cascaded control flow, the optimal control (LQI \cite{LQI0}) is applied. LQI implements integral control to compensate for unstructured but fixed uncertainties, thereby helping enhance the robustness, especially under singular forms. We also augment the original LQI framework in \cite{hydrus-ar2016} by adding an additional cost term.}

\revise{We first derive the rotational dynamics of \equref{eq:rotational_dynamics2} in the manner of the original LQI framework as follows:
\begin{eqnarray}
\label{eq:rot_state_equation}
& \dot{\bm{x}} = A \bm{x} + B \bm{\lambda} + D (I_{\Sigma}^{-1}\bm{\omega} \times I_{\Sigma} \bm{\omega}), \\
& A \in \mathcal{R}^{9 \times 9}, \hspace{1mm}
  B \in \mathcal{R}^{9 \times 4}, \hspace{1mm}
  D \in \mathcal{R}^{9 \times 3}, \nonumber \\
  &  \bm{x} = \left[e_x \ \dot{e}_x \ e_y \ \dot{e}_y \ e_z \ \dot{e}_z \ \int e_x  \ \int e_y  \ \int e_z \right]^{\mathrm{T}} \in \mathcal{R}^9, \nonumber \\
&  \left[e_x \ e_y \ e_z \right]^{\mathrm{T}} = {}^{\scriptframe{W}}\bm{\alpha}^{\mathrm{des}}_{\scriptframe{CoG}} - {}^{\scriptframe{W}}\bm{\alpha}_{\scriptframe{CoG}} \nonumber \\
&  \left[\dot{e}_x \ \dot{e}_y \ \dot{e}_z \right]^{\mathrm{T}} =  {}^{\scriptframe{CoG}}\bm{\omega}^{\mathrm{des}}_{\scriptframe{CoG}}  - {}^{\scriptframe{CoG}}\bm{\omega}_{\scriptframe{CoG}}. \nonumber
\end{eqnarray}
$A$ and $D$ are sparse matrices, where only $A_{12}, A_{34}, A_{56}, A_{71}, A_{83}, A_{95}, D_{21}, D_{42}$, and  $D_{63}$ are 1, and $B = \left[\begin{array}{ccccccc} {\bf 0}_{4\times 1} & -B_1 & {\bf 0}_{4\times 1} & -B_2 & {\bf 0}_{4\times 1} & -B_3  & {\bf 0}_{4\times 3}  \end{array}\right]^{\mathrm{T}}$, $\left[\begin{array}{ccc} B_1 &  B_2 & B_3\end{array}\right]^{\mathrm{T}} =  I_{\Sigma}^{-1} Q_{\mathrm{r}}(\bm{q}, \bm{\psi})$.
}

\revise{
Regarding the cost function for the state equation \equref{eq:rot_state_equation}, we extend our previous work \cite{hydrus-ar2016}, which has an additional term for suppressing the translational force, particularly the lateral force, caused by the attitude control:}
\begin{eqnarray}
\label{eq:lqi-cost-func}
&J = \int^{\infty}_0 \left(\bar{\bm{x}}^{\mathrm{T}} M  \bar{\bm{x}} + \bm{\lambda}^{\mathrm{T}} N \bm{\lambda} \right)dt, \\
\label{eq:lqi-input-cost}
&N = W_1 +  Q_{\mathrm{t}}^{\mathrm{T}}(\bm{q}, \bm{\psi}) W_2 Q_{\mathrm{t}}(\bm{q}, \bm{\psi}),
\end{eqnarray}
where the second term in \equref{eq:lqi-input-cost} corresponds to the minimization of the norm of the force generated by the attitude control, since $\|{}^{\scriptframe{CoG}}\bm{f}\|^2 = {}^{\scriptframe{CoG}}\bm{f}^{\mathrm{T}}{}^{\scriptframe{CoG}}\bm{f} = \bm{\lambda}^{\mathrm{T}} Q_{\mathrm{t}}(\bm{q}, \bm{\psi}) ^{\mathrm{T}} Q_{\mathrm{t}}(\bm{q}, \bm{\psi})  \bm{\lambda}$ \revise{from \equref{eq:total_wrench1}}. The diagonal weight matrices $M$, $W_1$, and $W_2$ balance the performance of the convergence to the desired state, suppression of the control input, and suppression of the translational force generated by the attitude control.

Then, a feedback gain matrix $K_x$ can be obtained by solving the related algebraic Riccati equation (ARE) driven by the state equation  \equref{eq:rot_state_equation} and the cost function \equref{eq:lqi-cost-func}. Finally, the desired control input regarding the attitude control $\bm{\lambda}^{\mathrm{des}}_{\mathrm{att}}$ can be given by:
\begin{eqnarray}
  \label{eq:desired_att_thrust_force}
  \bm{\lambda}^{\mathrm{des}}_{\mathrm{att}} = K_x\bm{x} +\revise{ Q^{\#}_{\mathrm{r}}(\bm{q}, \bm{\psi})  \bm{\omega} \times I_{\Sigma} \bm{\omega}},
\end{eqnarray}
\revise{where $(\cdot)^{\#}$ denotes the MP-pseudo-inverse of a full-rank matrix.}

\subsection{Position Control}
The position control is developed based on the method presented in \cite{lee-quadrotor-se3-control}, which first calculates the desired total force based on the common PID control, then converts the desired total force to the desired thrust vector and the \revise{desired} roll and pitch angles.
The desired total force can be given by:
\begin{eqnarray}
  \label{eq:pid_pos}
     {\bm f}^{\mathrm{des}} = m_{\Sigma} (K_P \bm{e}_r + K_I \int \bm{e}_{r} dt + K_D \dot{\bm{e}}_{r} + \ddot{\bm r}^{\mathrm{des}}),
\end{eqnarray}
where $\bm{e}_r =  {}^{\scriptframe{W}}\bm r_{\scriptframe{CoG}}^{\mathrm{des}} - {}^{\scriptframe{W}}\bm r_{\scriptframe{CoG}}$.

Then the desired roll and pitch angles can be given by:
\begin{eqnarray}
  \label{eq:desired_roll}
  &{}^{\scriptframe{W}}\alpha_{\scriptframe{CoG}_x}^{\mathrm{des}} = atan^{-1}(-\bar{f}_y, \sqrt{\bar{f}^2_x + \bar{f}^2_z}), \\
  \label{eq:desired_pitch}
  &{}^{\scriptframe{W}}\alpha_{\scriptframe{CoG}_y}^{\mathrm{des}} = atan^{-1}(\bar{f}_x, \bar{f}_z), \\
  & \left[\begin{array}{ccc}f_x & f_y & f_z \end{array}\right]^{\mathrm{T}}   = R_z^{-1}({}^{\scriptframe{W}}\alpha_{\scriptframe{CoG}_z}) {\bm f}^{\mathrm{des}}, \nonumber
\end{eqnarray}
\revise{where $R_{Z}(\cdot)$ is a rotation matrix that only rotates along the $z$ axis with a certain angle.} ${}^{\scriptframe{W}}\alpha_{\scriptframe{CoG}_x}^{\mathrm{des}}$ and ${}^{\scriptframe{W}}\alpha_{\scriptframe{CoG}_y}^{\mathrm{des}}$ are then transmitted to the attitude controller as shown in \figref{figure:control_diagram}.

On the other hand,  the desired collective thrust force can be calculated as follows:
\begin{eqnarray}
  \label{eq:collective thrust}
  f_T^{\mathrm{des}} = ({}^{\scriptframe{W}}R_{\scriptframe{CoG}} \bm{b}_3)^{\mathrm{T}} {\bm f}^{\mathrm{des}} ,
\end{eqnarray}
where  $\bm{b}_3$ is a unit vector $\left[\begin{array}{ccc}0 & 0& 1 \end{array} \right]^{\mathrm{T}}$, and ${}^{\scriptframe{W}}R_{\scriptframe{CoG}}$ is the orientation of the frame $\{CoG\}$.

Using \eqref{eq:collective thrust}, the allocation from the collective thrust force to the thrust force vector can be given by:
\begin{eqnarray}
  \label{eq:collective_thrust2}
  \bm{\lambda}^{\mathrm{des}}_{\mathrm{pos}} = \frac{\bm{\lambda}_s}{m_{\sum} g} f_T^{\mathrm{des}},
\end{eqnarray}
where \revise{$\bm{\lambda}_s$ is the static thrust force vector  as expressed by \equref{eq:lambda_s},} which only balances with the gravity force and thus does not affect the rotational dynamics. Eventually, the final \revise{desired} thrust force for each rotor should be the sum of outputs from the attitude and position control: $\bm{\lambda}^{\mathrm{des}} = \bm{\lambda}^{\mathrm{des}}_{\mathrm{att}} + \bm{\lambda}^{\mathrm{des}}_{\mathrm{pos}}$.

\subsection{Stability}
\revise{Given that the whole control framework is based on the near-hover assumption}, the desired roll and pitch angles sent to the attitude control in the hovering task would converge to zero under ideal conditions. This implies that the output of the attitude control $\bm{\lambda}^{\mathrm{des}}_{\mathrm{att}}$ and the lateral force generated by the attitude control $Q_{\mathrm{t}}(\bm{q}, \bm{\psi}) \bm{\lambda}^{\mathrm{des}}_{\mathrm{att}} $ would converge to zero, and $\bm{\lambda}^{\mathrm{des}}_{\mathrm{pos}}$ would converge to $m_{\sum} \bm{g}$ . Therefore, convergence to a fixed position can be guaranteed.
Furthermore, the integral control in both position and attitude control can compensate for the unstructured but fixed uncertainties other than gravity.

\section{Configuration Planning}
\label{sec:planning}

\subsection{Singular Form}
According to \cite{hydrus-ijrr2018}, there are several types of singular \revise{forms} in multilinked models without tilting  propellers (i.e., $\beta = 0$).
In the case of quad-types, there are two types of singular \revise{forms}, and they are shown in \figref{figure:abst}(A) and \figref{figure:abst}(B).

The group of singular forms like \figref{figure:abst}(A) can be expressed as:
\begin{equation}
  \label{eq:singular1}
    \mathcal{S}_1  :=  \{ \bm{q} \in \mathcal{R}^3  \mid q_{1} = - q_{3} = \revise{q},\hspace{2mm} q_2 = 0 ; -\frac{\pi}{2} \leq \revise{q} \leq  \frac{\pi}{2} \}.
  \end{equation}
Under such forms, \revise{$Q_{\mathrm{r}}(\bm{q}, \bm{\psi})$ in \equref{eq:rotational_dynamics2} cannot be a full-rank matrix if $\beta = 0$}.

The second group of singular forms like \figref{figure:abst}(B) can be expressed as:
\begin{equation}
  \label{eq:singular2}
    \mathcal{S}_2 := \{\bm{q} \in \mathcal{R}^3  \mid q_{1} =-q_{2} = q_3; -\frac{\pi}{2} \leq q_1 \leq  \frac{\pi}{2} \}.
\end{equation}
Under such \revise{forms}, all rotors are aligned along one straight line, and controllability around this line is entirely lost if $\beta = 0$.

\subsection{Rotational Controllability}

Controllability regarding rotational motion is an important factor for checking the validity under a form with certain joint angles $\bm{q}$. A quantity called feasible control torque convex ${\mathcal V}_{T}(\bm{q}, \bm{\psi})$ was introduced by \cite{odar-tasme2018} to validate rotational controllability in all axes, which is given by:
\begin{align}
  \label{eq:feasible_control_torque_convex}
  &{\mathcal V}_T(\bm{q}, \bm{\psi}) := \nonumber \\
  &\{ {}^{\{C\}}\bm{\tau}(\bm{q}, \bm{\psi}) \in {\mathcal R}^3 | {}^{\{C\}}\bm{\tau}(\bm{q}, \bm{\psi}) = \displaystyle \sum_{i=1}^{N} \lambda_i \bm{v}_i(\bm{q}, \bm{\psi}), 0 \leq \lambda_i \leq \lambda_{\mathrm{max}} \}, 
\end{align}
where $\bm{v}_i(\bm{q}, \bm{\psi}) = ({}^{\{C\}}\hat{\bm{p}}_{\{F_i\}}(\bm{q}, \bm{\psi}) + \kappa_i E_{3 \times 3} ) {}^{\{C\}}R_{\{F_i\}}(\bm{q}, \bm{\psi}) \bm{b}_z$ according to \equref{eq:wrench_torque}.

Inside the convex ${\mathcal V}_T(\bm{q}, \bm{\psi})$, the guaranteed minimum control torque $\tau_{\mathrm{min}}(\bm{q}, \bm{\psi})$ is further introduced, which has following property:
\begin{eqnarray}
  \label{eq:guaranteed_minimum_control_torque}
  \|{}^{\{C\}}\bm{\tau}(\bm{q}, \bm{\psi}) \| \leq \tau_{\mathrm{min}}(\bm{q}, \bm{\psi}) \Rightarrow {}^{\{C\}}\bm{\tau}(\bm{q}, \bm{\psi}) \in {\mathcal V}_T(\bm{q}, \bm{\psi}),
\end{eqnarray}

Using the distance $d^{\tau}_{ij}(\bm{q}, \bm{\psi})$ which is from the origin \revise{$\left[0 \ 0\ 0 \right]^{\mathrm{T}}$} to a plane of convex ${\mathcal V}_T(\bm{q}, \bm{\psi})$ along its normal vector $\frac{\bm{v}_i \times \bm{v}_j}{\|\bm{v}_i \times \bm{v}_j\|}$, $\tau_{\mathrm{min}}(\bm{q}, \bm{\psi})$ can be given by:
\begin{eqnarray}
  \label{eq:feasible_control_fxy}
  &d^{\tau}_{ij}(\bm{q}, \bm{\psi}) = \displaystyle \sum_{k=1}^{N} \max (0, \lambda_{\mathrm{max}} \frac{(\bm{v}_i \times \bm{v}_j)^{\mathrm{T}}}{\|\bm{v}_i \times \bm{v}_j\|} \bm{v}_k), \\
  \label{eq:feasible_control_fxy}
  &{\tau}_{\mathrm{min}}(\bm{q}, \bm{\psi}) = \displaystyle \min_{i,j \in {\mathcal I}} d^{\tau}_{ij}(\bm{q}, \bm{\psi}),
\end{eqnarray}
where $\mathcal{I} := \{1, 2, \cdots, N\}$.
Evidently, ${\tau}_{\mathrm{min}}(\bm{q}, \bm{\psi}) > 0$ is the condition to guarantee  rotational controllability in all axes. However, $\tau_{\mathrm{min}}$ in the singular groups $\mathcal{S}_1$ and  $\mathcal{S}_2$ would shrink to zero if $\beta = 0$.

\subsection{Optimization of Vectoring Angles}
If $\beta > 0$, singular forms can be resolved by proper vectoring angles $\bm{\psi}$ ensuring  $\tau_{\mathrm{min}}(\bm{q}) > 0$.
In addition to maximizing controllability regarding the rotational motion, it is also important to minimize the collective thrust force which corresponds to the energy efficiency, as well as the gap among thrust forces which corresponds to the maximization of the control margin. Then, an optimization problem with an integrated cost function is designed as follows:
\begin{equation}
  \label{eq:optimization}
\displaystyle \max_{\bm{\psi}} (w_1 \tau_{\mathrm{min}}(\bm{q}, \bm{\psi}) + \frac{w_2}{\|\bm{\lambda}_s(\bm{q}, \bm{\psi}) \|} + \frac{w_3}{\sigma^2(\bm{\lambda}_s(\bm{q}, \bm{\psi}))}),
\end{equation}
\begin{eqnarray}
  \label{eq:contriants1}
  \mathrm{s.t.} \hspace{3mm} & \alpha_{\mathrm{min}} \leq \alpha_x (\bm{q}, \bm{\psi}) \leq \alpha_{\mathrm{max}}, \\
  \label{eq:contriants2}
  &\alpha_{\mathrm{min}} \leq \alpha_y (\bm{q}, \bm{\psi}) \leq \alpha_{\mathrm{max}},
\end{eqnarray}
where $\|\bm{\lambda}_s(\bm{q}, \bm{\psi}) \|$ and $\sigma^2(\bm{\lambda}_s(\bm{q}, \bm{\psi}))$ are the norm and variance, respectively, of the hovering thrust vector obtained from \equref{eq:lambda_s}. Given that the flight is assumed to be near-hovering, the hovering thrust is chosen. $w_1, w_2$, and $w_3$ are the positive weights to balance between the rotational controllability, energy efficiency, and control margin.
On the other hand, $\alpha_x (\bm{q}, \bm{\psi}), \alpha_y (\bm{q}, \bm{\psi})$  in \equref{eq:contriants1} and \equref{eq:contriants2} are the roll and pitch angles between the frames $\{C\}$ and $\{CoG\}$, and they are obtained from \equref{eq:R_CoG_roll} and \equref{eq:R_CoG_pitch}. \revise{These constraints are introduced to suppress the difference between  ${}^{\scriptframe{CoG}}R_{\scriptframe{C}}$ in \equref{eq:R_CoG2}  and the identity matrix $E_{3 \times 3}$, thereby making the multilinked model as level as possible in the hovering state.}

\begin{figure}[b]
  \begin{center}
    \includegraphics[width=\columnwidth]{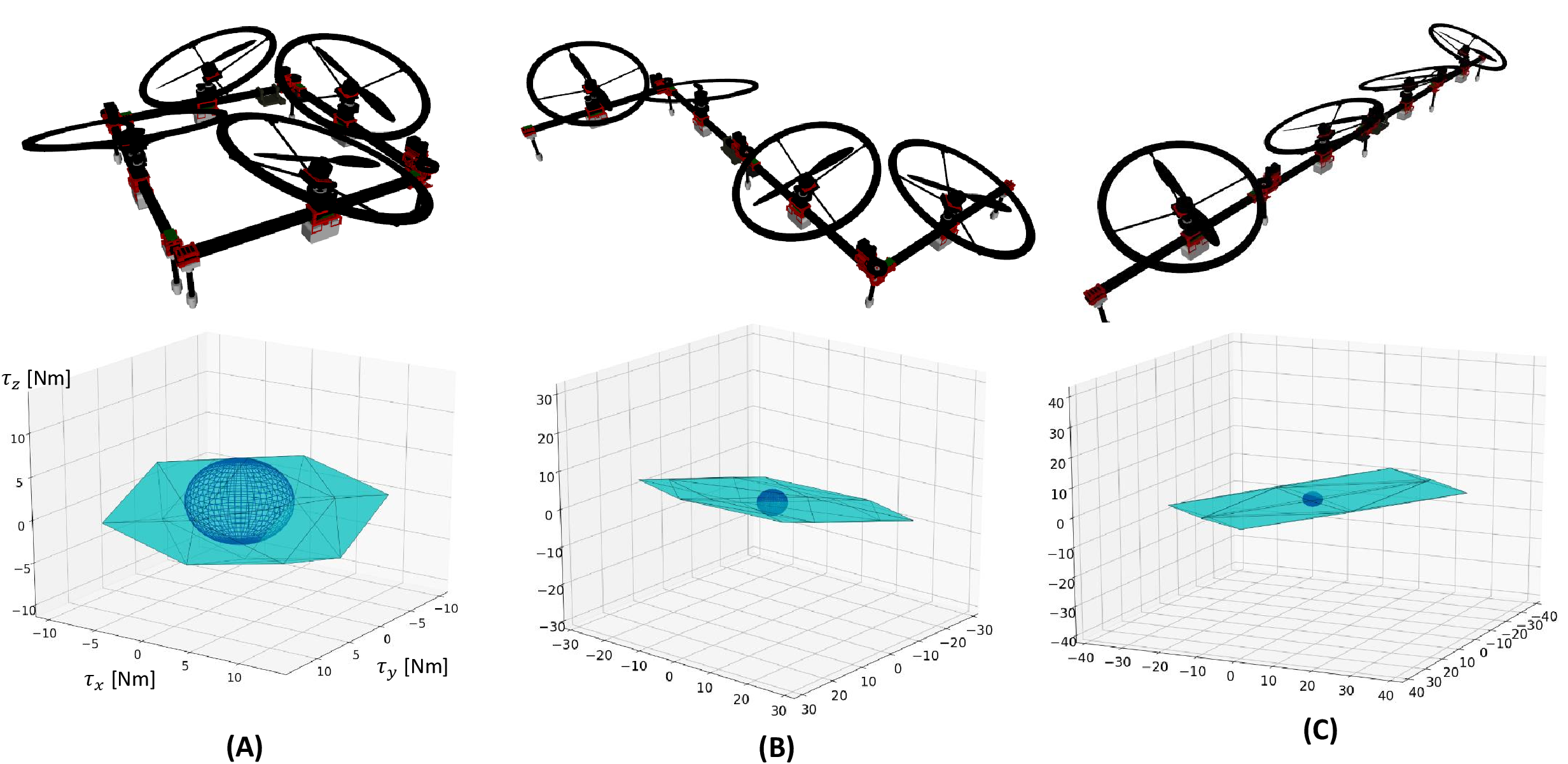}
    \vspace{-5mm}
    \caption{Optimization results in case of a quad-type model as shown in \figref{figure:platform}. Light blue convex denotes the feasible control torque convex ${\mathcal V}_T$, and the radius of the dark blue sphere denotes the guaranteed minimum control torque $\tau_{\mathrm{min}}$. {\bf (A)} The optimal vectoring angles under the normal form (i.e., $q_i = \frac{\pi}{2}$ rad) are  $\bar{\bm{\psi}} = \left[3.14 \ 0.12 \ 3.09 \ 0.00 \right]^{\mathrm{T}}$ rad, and $\tau_{\mathrm{min}}$ is 4.81 Nm; {\bf (B)} The optimal vectoring angles under the form in the singular group $\mathcal{S}_1$ (i.e., $\bm{q} = \left[-\frac{\pi}{2} \ 0 \ \frac{\pi}{2} \right]^{\mathrm{T}}$ rad) are  $\bar{\bm{\psi}} = \left[-0.69 \ -1.82 \ 2.73 \ -1.52 \right]^{\mathrm{T}}$ rad, and $\tau_{\mathrm{min}}$ is 3.28 Nm; {\bf (C)} The optimal vectoring angles  under the form in the singular group $\mathcal{S}_2$ (i.e., $q_i = 0$ rad) are  $\bar{\bm{\psi}} = \left[1.39 \ -1.23 \ -1.77 \ 1.77 \right]^{\mathrm{T}}$ rad, and $\tau_{\mathrm{min}}$ is 2.38 Nm.}
    \vspace{-5mm}
     \label{figure:fc_t_min}
  \end{center}
\end{figure}

\figref{figure:fc_t_min} shows several results of optimization in case of a quad-type model as shown in \figref{figure:platform}. 
The guaranteed minimum control torque (i.e., radius of the blue sphere) is sufficiently large even under singular forms such as those in \figref{figure:fc_t_min}(B) and \figref{figure:fc_t_min}(C).

\subsection{Continuity of Vectoring Angles During Deformation}
The optimization problem presented in \equref{eq:optimization} $\sim$ \equref{eq:contriants2} cannot guarantee the continuity of the change in the vectoring angles during deformation. 
Thus, the following additional temporal constraints regarding the vectoring angles are added to the optimization problem:
\begin{eqnarray}
  \label{eq:contriants3}
  \bar{\psi}_i(t-1) - \delta \psi \leq \psi_i(t) \leq \bar{\psi}_i(t-1) + \delta \psi, \hspace{2mm}(1 \leq i \leq N),
\end{eqnarray}
where $\delta \psi$ is a small constant value, and $\bar{\bm{\psi}}(t-1) = \left[\bar{\psi}_1(t-1) \ \bar{\psi}_2(t-1) \ \cdots \ \bar{\psi}_N(t-1)  \right]^{\mathrm{T}}$ denotes the optimal vectoring angles at $t-1$.
\revise{These constraints are not applied for the initial form (i.e., $t = 0$) with the aim of performing a global search for the initial form.}

\begin{figure}[t]
  \begin{center}
    \includegraphics[width=\columnwidth]{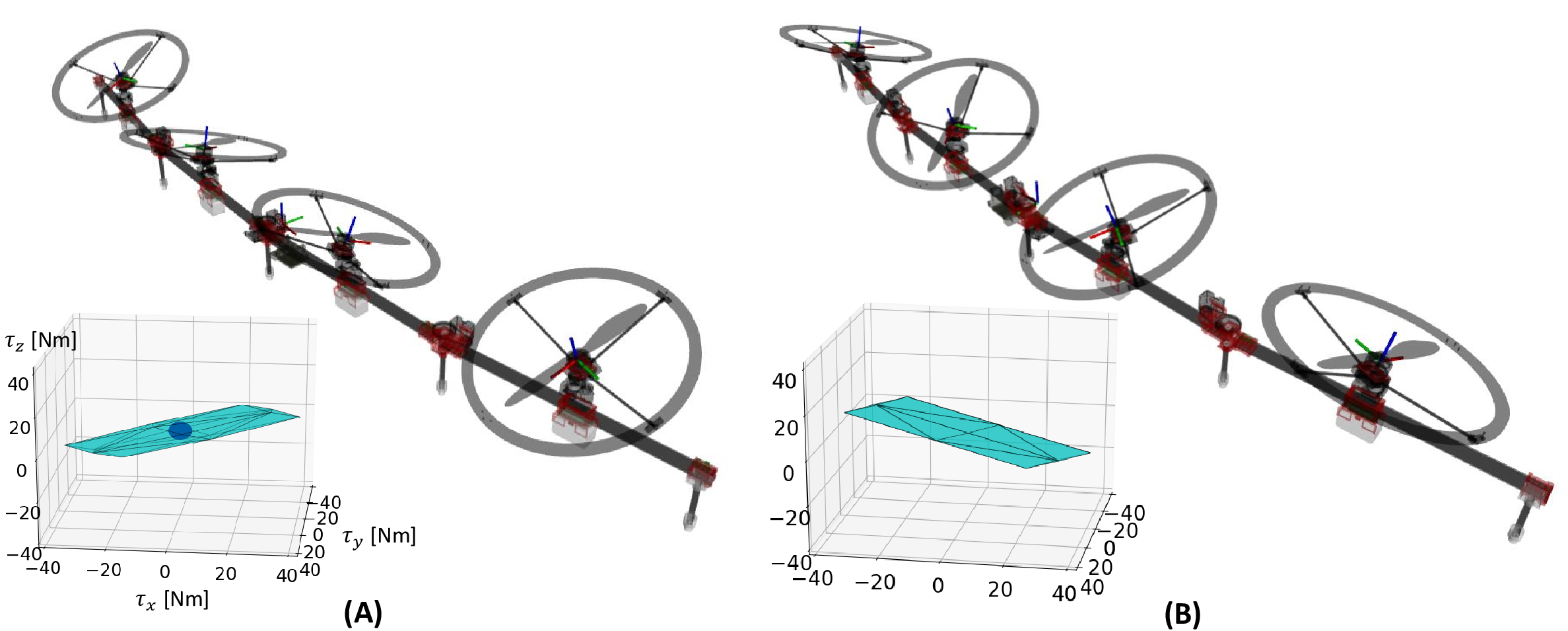}
    \vspace{-5mm}
    \caption{\revise{Corner case ($q_i = 0.11$ rad) that does not have dual solutions for vectoring angles. {\bf (A)} The result  ($\bar{\bm{\psi}} = \left[1.88\ -1.04\ -1.93 \ 1.41 \right]^{\mathrm{T}}$ rad) obtained from \equref{eq:optimization} $\sim$ \equref{eq:contriants3};  the guaranteed minimum control torque is 3.85 Nm (i.e., the blue sphere in the subplot). {\bf (B)} The form with the same joint angles and reversed (dual) vectoring angles (i.e., $\psi_i = \bar{\psi} + \pi$). The guaranteed minimum control torque is zero (i.e., no blue sphere in the plot).}}
     \label{figure:dual_solution_corner_case}
  \end{center}
\end{figure}

For most of joint angles $\bm{q}$, there are dual sets of vectoring angles that have a similar performance: a primal solution $\bar{\bm{\psi}}(\bm{q})$ calculated from \equref{eq:optimization} $\sim$ \equref{eq:contriants3}, and a dual solution \revise{ $\bar{\bm{\psi}}(\bm{q})^{'} \approx \left[\bar{\psi}_1(\bm{q}) + \pi \ \bar{\psi}_2(\bm{q})  + \pi \ \cdots \ \bar{\psi}_N(\bm{q})  + \pi \right]^{\mathrm{T}}$} indicating almost opposite directions in each vectoring apparatus. \revise{However, there exist corner cases like \figref{figure:dual_solution_corner_case}, where the guaranteed minimum control torque shrinks to zero with the reversed vectoring angles $\bar{\bm{\psi}}(\bm{q})^{'}$.}
Such a set of joint angles is defined as $\bm{q}_{\mathrm{corner}}$.
The following are important properties of $\bm{q}_{\mathrm{corner}}$:  a) The opposite set, namely, $-\bm{q}_{\mathrm{corner}}$, also belongs to the corner case. b) $\bar{\bm{\psi}}(\bm{q}_{\mathrm{corner}})$ is continuous with the invalid solution $\bar{\bm{\psi}}^{'}(-\bm{q}_{\mathrm{corner}})$, but is not continuous with the valid solution $\bar{\bm{\psi}}(-\bm{q}_{\mathrm{corner}})$.
Thus, linear transition from $\bm{q}_{\mathrm{corner}}$ to $-\bm{q}_{\mathrm{corner}}$ (\revise{$q_i(t) = q_{\mathrm{corner}_i} - \dot{q} t; t \in \left[0, \  \frac{2q_{\mathrm{corner}_i}}{\dot{q}}\right]$} ) is impossible. This also implies an impossible deformation involving a wider linear transition from the form of \figref{figure:fc_t_min}(A) to the opposite form. 
Nevertheless, this opposite form can still be arrived from the form  of \figref{figure:fc_t_min}(A) via roundabout paths (e.g., passing through the form of \figref{figure:fc_t_min}(B)).
Furthermore, the linear transition (i.e., \figref{figure:fc_t_min}(A) $\rightarrow$ \figref{figure:fc_t_min}(C)) is also possible, since one of the two solutions of vectoring angles  in \figref{figure:fc_t_min}(A) is continuous with $\bar{\bm{\psi}}(\bm{q}_{\mathrm{corner}})$ where the robot must pass during the transition.

\section{Experiments}
\label{sec:experiment}

\subsection{Robot Platform}
\begin{figure}[!t]
  \begin{center}
    \includegraphics[width=0.9\columnwidth]{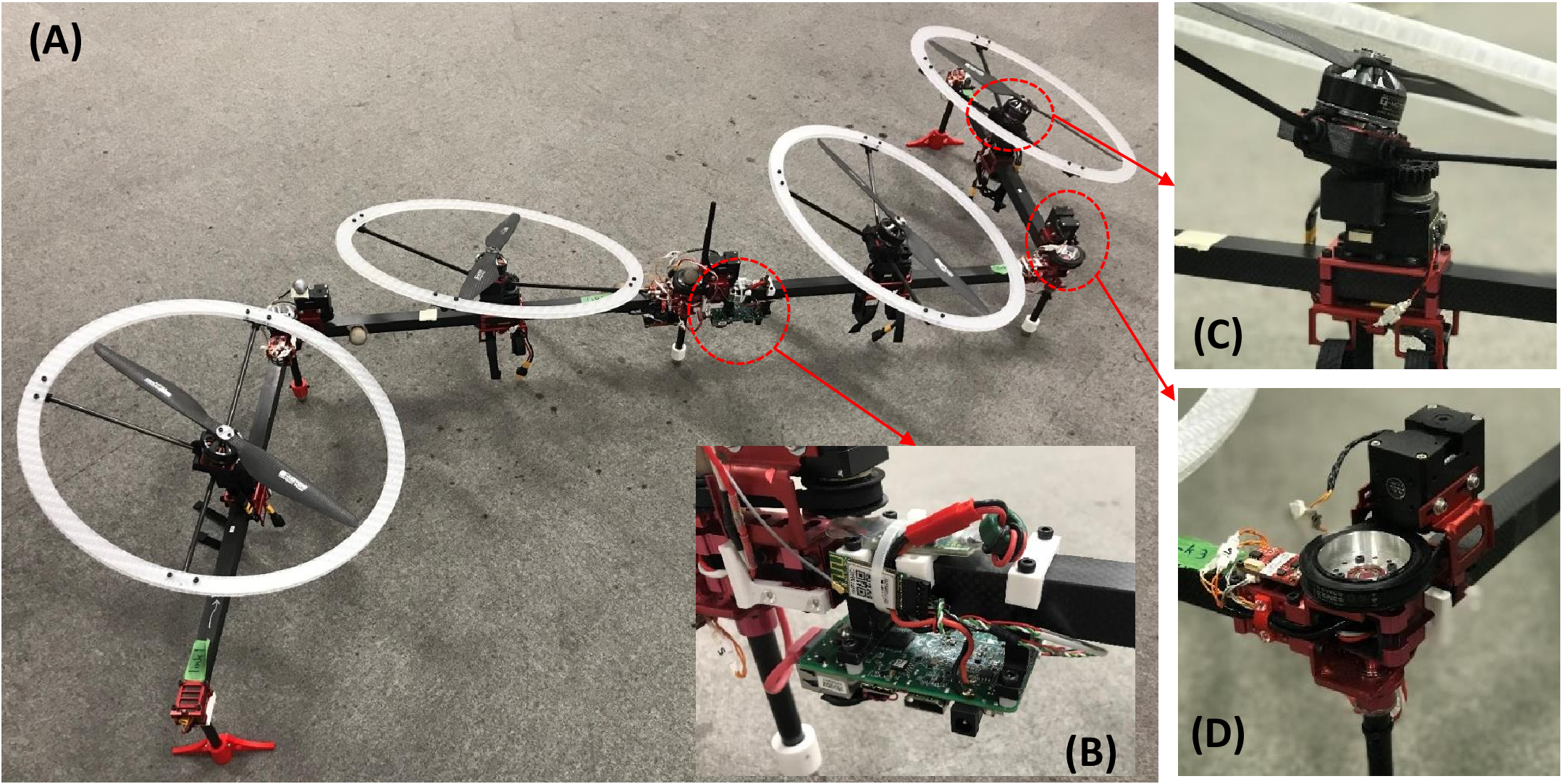}
    \vspace{-3mm}
    \caption{{\bf A}: Quad-type multilinked platform used in this work. {\bf B}: Onboard computer (AAEON UP-BOARD with Intel Atom x5-Z8350 quad core CPU) to process the flight control and planning for  vectoring angles. {\bf C}: Propeller vectoring apparatus actuated by a servo motor (Dynamixel MX28AR) under the propeller and rotor.  {\bf D}: Joint module actuated by a servo motor (Dynamixel MX28AR).}
    \vspace{-5mm}
    \label{figure:platform}
  \end{center}
\end{figure}

A quad-type multilinked robot platform is used in this work as shown in \figref{figure:platform}.
An onboard computer with an Intel Atom CPU is equipped to process the proposed flight control and planning for the vectoring angles. The whole weight is 4.7 kg, leading to a 10 min flight. Detailed specifications regarding the link module and the internal system can be found in our previous work \cite{anzai-hydrus-xi-iros2019}. In addition, a motion capture system is applied in our experiment to obtain the robot state (i.e., ${}^{\{W\}}{\bm r}_{\{L_1\}}, {}^{\{W\}}\dot{\bm r}_{\{L_1\}}, {}^{\{W\}}{R}_{\{L_1\}}, {}^{\{L_1\}}{\bm \omega}_{\{L_1\}}$).

\revise{The attitude control weight matrices for LQI in \equref{eq:lqi-cost-func} and \equref{eq:lqi-input-cost}  are $M = diag(1100, 80, 1100, 80, 100, 50, 10, 10, 0.5)$,  $W_1 = diag(1, 1, 1, 1)$, and $W_2 = diag(20, 20, 20)$, while the position control gains in \equref{eq:pid_pos} are  $K_{P} = diag(2.3, 2.3, 3.6)$, $K_I = diag(0.02, 0.02, 3.4)$, and  $K_D = diag(4.0, 4.0, 1.55)$.}

\revise{
For the nonlinear optimization problem \equref{eq:optimization} $\sim$ \equref{eq:contriants2}, the search space is four-dimensional in our quad-link type. A nonlinear optimization algorithm COBYLA \cite{COBYLA} is used with $w_1 = 1.0 $, $w_2 = 2.0, w_3 = 0.01, \alpha_{\mathrm{max}} = -\alpha_{\mathrm{min}} = 0.01$, and the average solving time in the onboard computer is approximately 0.05s (i.e., 20Hz). The joint velocity for the aerial deformation is set at 0.25 rad/s, implying that the maximum change in the joint angle in the planning interval is 0.0125 rad. The search range $\delta \psi$ in \equref{eq:contriants3} was set relatively large (i.e., 0.2 rad) so it can rapidly keep up with changes in the model configuration, especially around singular forms.
}

\subsection{Experimental Results}
\subsubsection{Circular Trajectory Tracking under Line-shape Form}
\begin{figure}[!h]
  \begin{center}
    \includegraphics[width=\columnwidth]{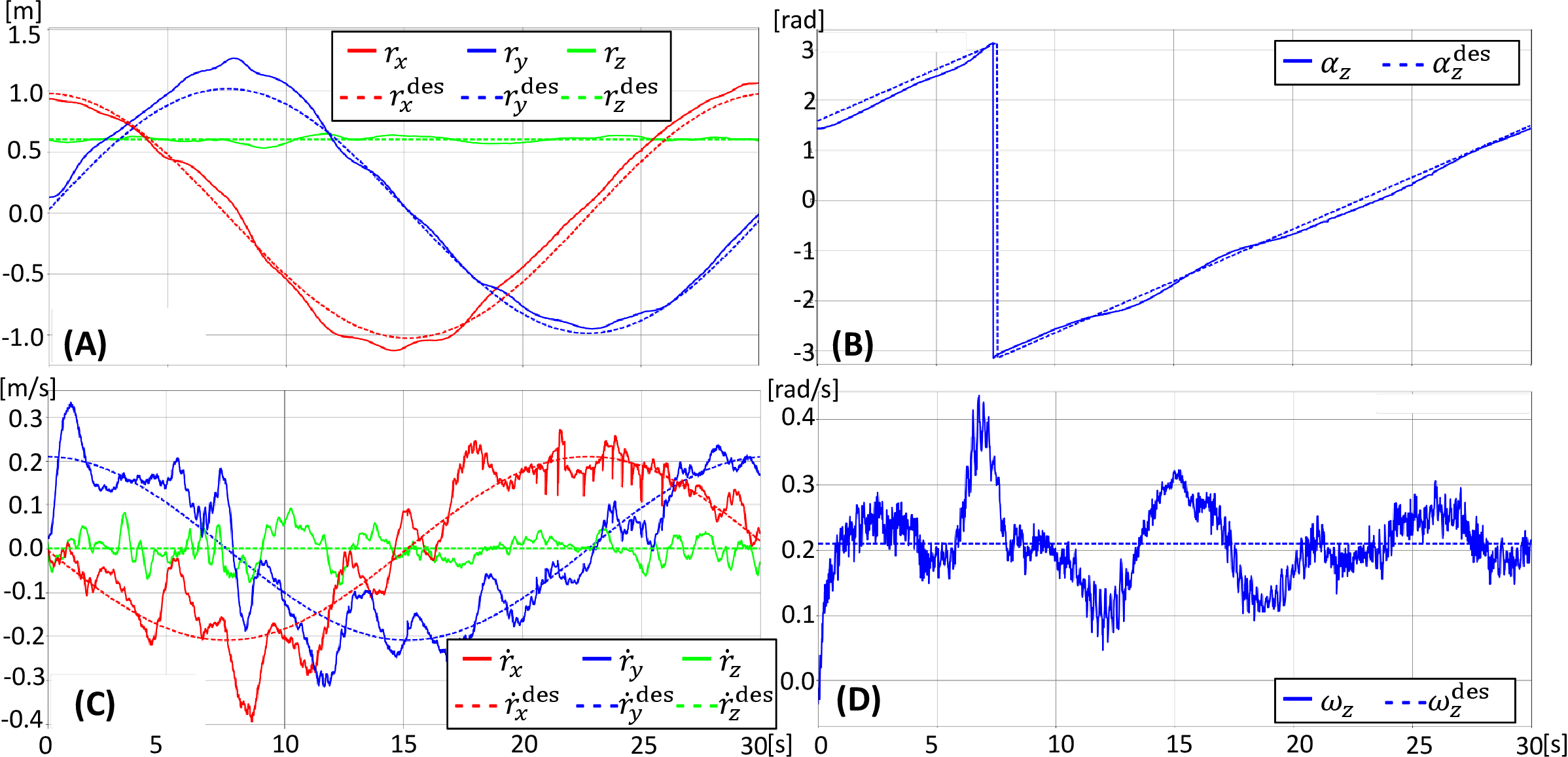}
    \vspace{-7mm}
    \caption{Plots of the motion performed to follow a circular trajectory in terms of position {\bf (A)}, yaw angle {\bf (B)}, linear velocity {\bf (C)}, and $z$  angular velocity {\bf (D)}. The motion can be found in our attached video.}
    \label{figure:circle_traj_plot}
  \end{center}
\end{figure}

\revise{
  We first evaluated the stability during trajectory tracking under the line-shape form of \figref{figure:abst}(B). We designed a 2D circular trajectory regarding the frame $\{CoG\}$ whose the radius is 1 m and period is 30 s, as shown in \figref{figure:circle_traj_plot}(A). Besides, we also designed the desired yaw angle to make the robot rotate $360^{\circ}$, as shown in \figref{figure:circle_traj_plot}(B). Tracking errors can be obtained by comparing the desired trajectory (dashed line) and the real trajectory (solid line) in each subplot of  \figref{figure:circle_traj_plot}. The maximum error in the lateral direction was 0.3 m at 8 s, as shown in \figref{figure:circle_traj_plot}(A), when the robot was very close to a wall, while the maximum error of the yaw angle was 0.22 rad at 1 s, as shown in \figref{figure:circle_traj_plot}(B).
  The RMS of the tracking errors regarding the 3D position and the yaw angle are [0.087, 0.094, 0.024] m and 0.112 rad, respectively.  The overall tracking stability is confirmed, demonstrating the effectiveness of the proposed control method not only for hovering but also for dynamic motion.
}

\subsubsection{Large-scale Deformation}
\begin{figure}[!b]
  \begin{center}
    \begin{minipage}{1.0\hsize}
      \begin{center}
        \includegraphics[width=0.95\columnwidth]{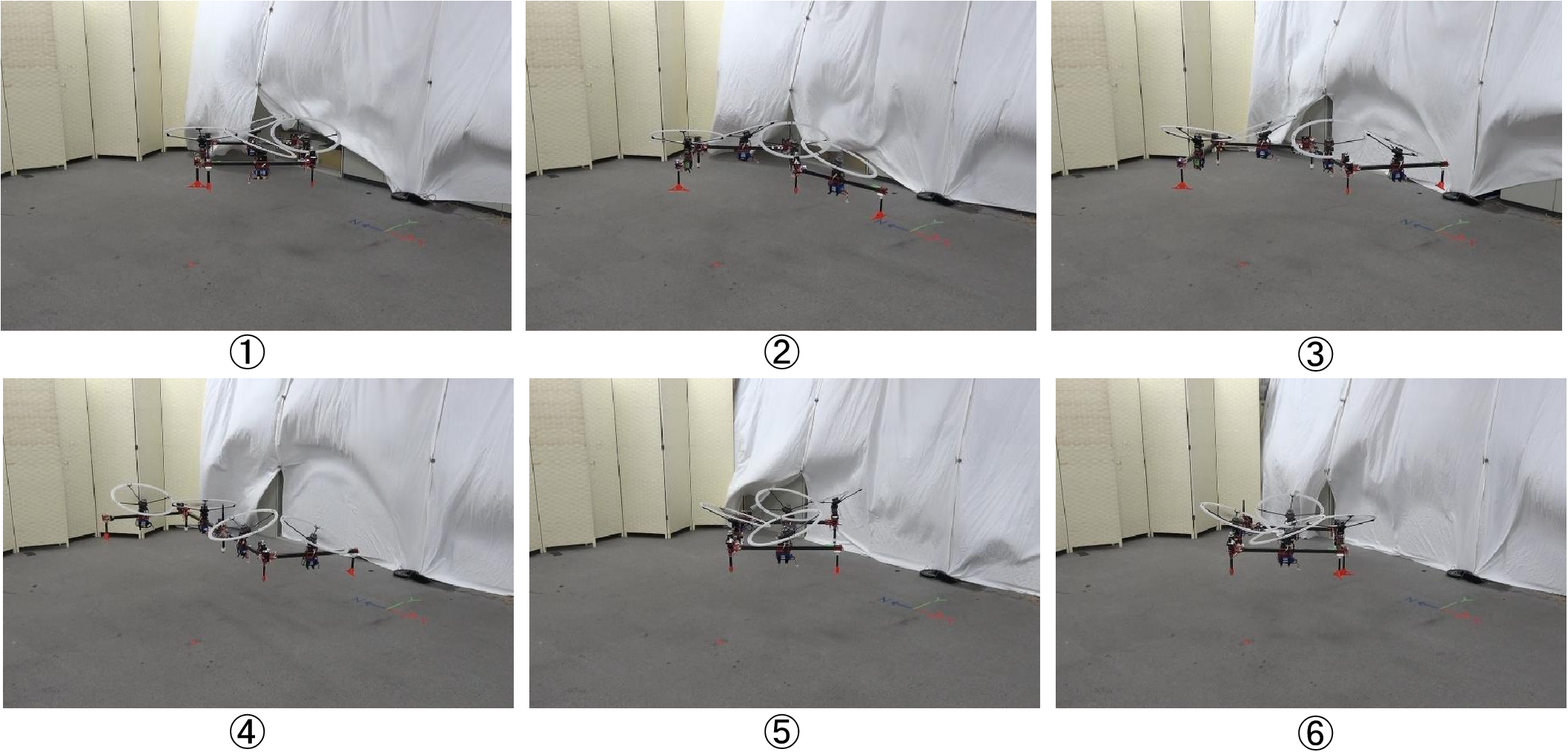}
        \vspace{-4mm}
        \caption{Snapshots of large-scale deformation where the numbers correspond to \figref{figure:full_deformation_plots}. The form in \textcircled{\scriptsize 4} is identical to that in \figref{figure:abst}(A).}
        \label{figure:full_deformation_images}
      \end{center}
    \end{minipage}
    \begin{minipage}{1.0\hsize}
      \begin{center}
        \includegraphics[width=0.95\columnwidth]{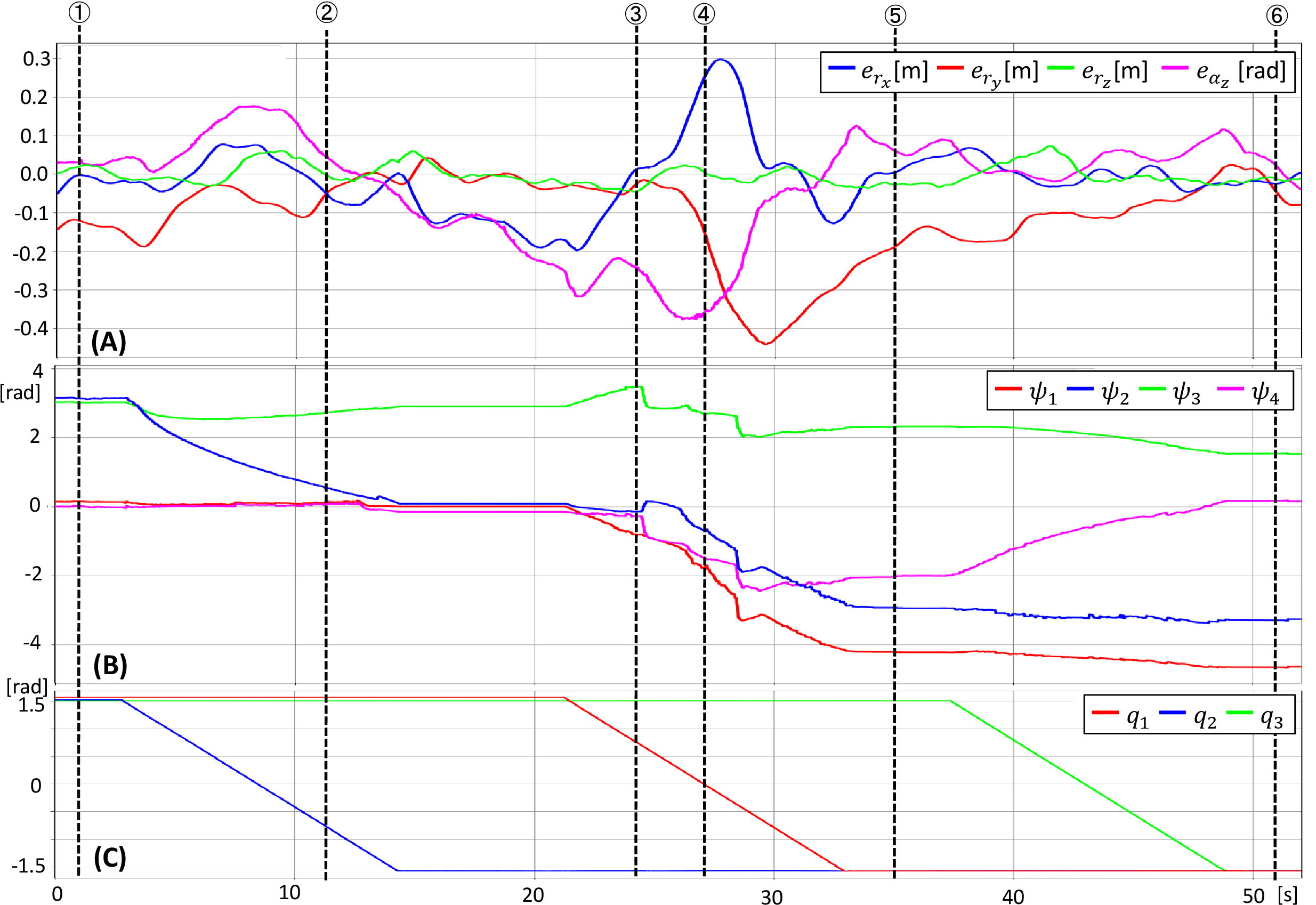}
        \vspace{-4mm}
        \caption{{\bf A}: Tracking errors regarding the position and yaw angle during the large-scale deformation of \figref{figure:full_deformation_images}. {\bf B}: Change in vectoring angles during deformation.  {\bf C}: Change in joint angles during deformation. The numbers at the top correspond to \figref{figure:full_deformation_images}.}
        \label{figure:full_deformation_plots}
      \end{center}
    \end{minipage}
  \end{center}
\end{figure}

We then evaluated the stability during a large-scale deformation as shown in \figref{figure:full_deformation_images}, where the joint angle linearly changed from $\frac{\pi}{2}$ to $-\frac{\pi}{2}$ one after another as shown in  \figref{figure:full_deformation_plots}(C).
The robot passed a singular form of \figref{figure:abst}(A) at \textcircled{\scriptsize 4}.
The maximum position errors  were $\left[ 0.3\ 0.45\ 0.07\right]$ m and the maximum yaw angle error was 0.38 rad as shown in \figref{figure:full_deformation_plots}(A), which increased significantly around \textcircled{\scriptsize 4}. We consider that the self-interference caused by the airflow from the tilting propellers acting on the other robot components became significantly larger during 25 s $\sim$ 33 s. If the robot is under a fixed form,  such a self-interference  would be constant; thus, can be compensated. However fluctuating interference during deformation cannot be easily handled by our proposed flight control.
Although slowing down the joint motion can be a straightforward solution, it is also possible to model such self-interference and consider it in the optimization problem of vectoring angles, which is an important future work.
Furthermore, the vectoring angles shown \revise{in} \figref{figure:full_deformation_plots}(B) changed continuously, demonstrating the feasibility of our proposed planning method.

\subsubsection{Deformation to the Line-shape Form}
\begin{figure}[t]
  \begin{center}
    \begin{minipage}{1.0\hsize}
      \begin{center}
        \includegraphics[width=0.9\columnwidth]{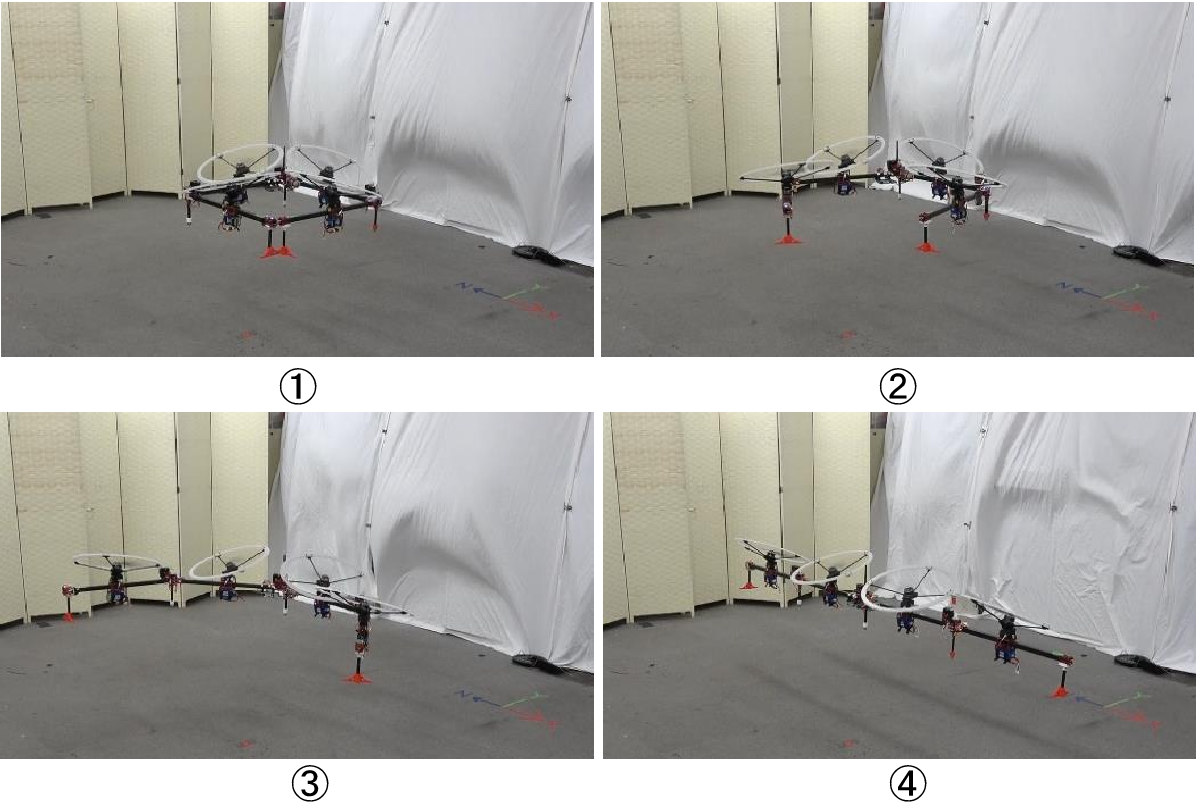}
        \vspace{-4mm}
        \caption{Snapshots of the deformation from a normal form to a line-shape form, where the numbers correspond to \figref{figure:normal_to_line_plots}.}
        \label{figure:normal_to_line_images}
      \end{center}
    \end{minipage}
    \begin{minipage}{1.0\hsize}
      \begin{center}
        \includegraphics[width=0.95\columnwidth]{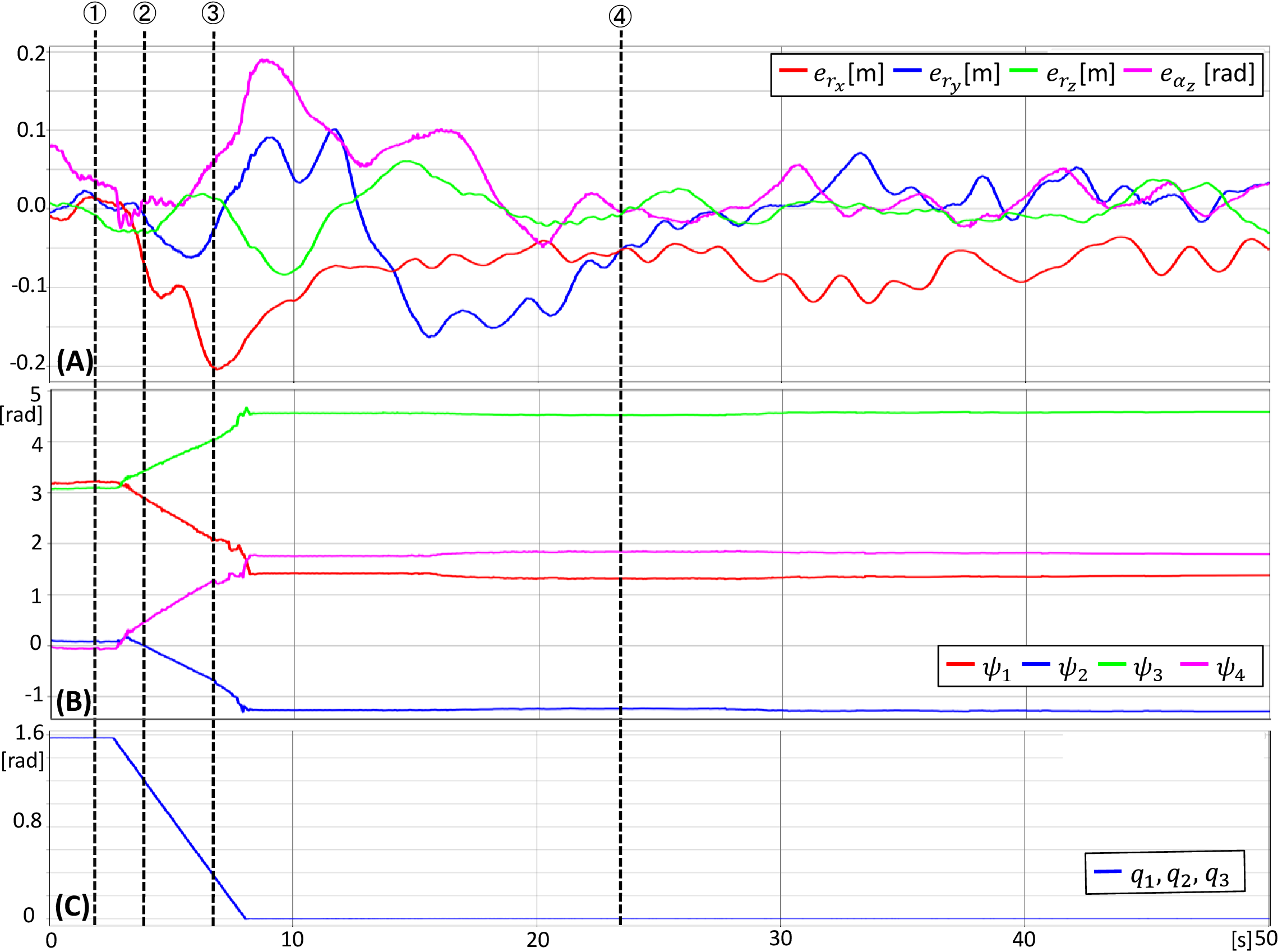}
        \vspace{-4mm}
        \caption{{\bf A}: Tracking errors regarding the position and yaw angle during deformation of \figref{figure:normal_to_line_images}. {\bf B}: The change in the vectoring angles during deformation. {\bf C}: The change in the joint angles during deformation. The numbers at the top correspond to \figref{figure:normal_to_line_images}.}
        \label{figure:normal_to_line_plots}
      \end{center}
    \end{minipage}
  \end{center}
\end{figure}

We also evaluated the stability during deformation to the line-shape form, as shown in \figref{figure:normal_to_line_images}. \revise{The tracking errors shown in \figref{figure:normal_to_line_plots}(A) were smaller than the tracking errors during deformation in \figref{figure:full_deformation_images}, where the largest deviation in the lateral direction was -0.2 m at \textcircled{\scriptsize 3}.}
\revise{After the deformation, the robot kept hovering under the line-shape form (i.e., $8 s \sim$). The small deviation in all axes indicates convergence to the desired fixed position and yaw angle, which demonstrates the robustness provided by our proposed control method.}
On the other hand,  \figref{figure:normal_to_line_plots}(B) also demonstrates the continuous change in the vectoring angle. In summary, the feasibility of the proposed methods is successfully demonstrated through these experiments.

\section{Conclusions and Future Work}
\label{sec:conclusion}

\revise{
  In this paper, we present the achievement of singularity-free flight for a two-dimensional multilinked aerial robot by employing 1-DoF yaw-vectorable propellers. We extended the modeling and control method based on our previous work to address the net force with variable direction owing to the vectorable propeller.
  Then, an optimization-based planning method for the vectoring angles was developed to maximize the guaranteed minimum torque under arbitrary robot forms.
  The feasibility of the above methods is verified by experiments involving a quad-type model, including the trajectory tracking under the line-shape form, as well as the deformation passing such a challenging form.}

Maneuvering and manipulation involving singularity-free deformation will be performed in future work. On the other hand,
a crucial issue remaining in this work is the fluctuating self-interference caused by the airflow from the tilting propeller acting on the other parts of the robot, which influences stability during deformation.
An enhanced planning method for vectoring angles that can avoid such self-interference based on a kinematics model will, thus, be developed.

\bibliographystyle{junsrt}
\bibliography{main}

\end{document}